\documentclass[11pt]{article}
\usepackage[margin=1in]{geometry}
\usepackage[T1]{fontenc}
\usepackage{lmodern}
\usepackage{amsmath,amssymb}
\usepackage{booktabs,tabularx,multirow}
\usepackage{graphicx}
\usepackage[numbers,sort&compress]{natbib}
\usepackage[hidelinks]{hyperref}
\usepackage{xcolor}
\usepackage{caption}
\graphicspath{{./}}
\newcommand{\pub}{p50c}

\title{Writing Weights or Training Them?\\
Spatial Reasoning through Weight-Space Programming and Low-Rank Adaptation}
\author{Yalin Yang\textsuperscript{1,*}\quad Xiao Huang\textsuperscript{2,*}\quad Yanan Wu\textsuperscript{3}\quad Hao Yang\textsuperscript{4}\quad Ryan Zhenqi Zhou\textsuperscript{5}\\[1.5ex]
\small\textsuperscript{1}West Virginia GIS Technical Center, West Virginia University, Morgantown, WV 26505, USA\\
\small\textsuperscript{2}Department of Environmental Science, Emory University, Atlanta, GA, USA\\
\small\textsuperscript{3}Department of Geography, University of Central Arkansas, Conway, AR 72035, USA\\
\small\textsuperscript{4}Systems Engineering Program, Cornell University, Ithaca, NY 14853, USA\\
\small\textsuperscript{5}Yale Center for Geospatial Solutions, Yale University, New Haven, CT, USA}
\date{}

\begin{document}
\maketitle
\begingroup
\renewcommand{\thefootnote}{*}
\footnotetext[1]{Corresponding authors.\\
\textit{Email addresses:} \href{mailto:yalin.yang@mail.wvu.edu}{\nolinkurl{yalin.yang@mail.wvu.edu}} (Yalin Yang); \href{mailto:xiao.huang2@emory.edu}{\nolinkurl{xiao.huang2@emory.edu}} (Xiao Huang).}
\endgroup

\begin{abstract}
Weight-space programming lets a language model inspect and edit another model's weights, offering an alternative to gradient-based adaptation. We evaluated both approaches using Qwen3.5-0.8B on a fixed reference-frame task (128 items), StepGame (200 items), and two untrained distance-comparison tasks. With thinking enabled, the public patch (\pub) raised mean accuracy from 0.0\% to 22.1\% on the reference-frame task and from 19.2\% to 33.7\% on StepGame across three sampling seeds. Randomly permuting its weight change removed the gain. Further closed-form edits did not improve on \pub, whereas a rank-1 LoRA adapter trained on worked solutions reached 100\% and 98.8\%, respectively, with 0.09\% trainable parameters. Removing the worked solutions from supervision prevented the adapter from passing the learning gate. The trained adapters reached 82\% accuracy on unseen StepGame sentence templates, 42\% on held-out difficulty levels 6--10 after training on levels 1--5, and 33--50\% on unseen heading--answer combinations. Their target gains coincided with a sharp decline on the distance tasks: accuracy fell from 94\% and 92\% under \pub\ to 0--2\%, and their outputs applied the StepGame solution template to distance questions. With thinking disabled and reasoning requested in the visible answer, the base model also solved the distance tasks while remaining weak on the target tasks. The comparison reveals a trade-off between improving spatial task performance and preserving neighbouring skills: the public edit restored completion while preserving distance performance, whereas supervised adaptation produced much larger target gains together with interference across tasks.
\end{abstract}

\section{Introduction}

Large language models are increasingly adapted after pre-training through task-specific optimisation and targeted interventions. Parameter-efficient training fits a small set of parameters by gradient descent on task data. Parameter-efficient variants restrict the change to small adapters \citep{houlsby2019adapters}, continuous prompts \citep{li2021prefix,lester2021prompt} or low-rank weight updates \citep{hu2022lora,dettmers2023qlora,liu2024dora}, which makes adaptation cheap enough to be routine. Targeted interventions instead specify changes to weights or activations using measurements of the network, learned update rules or differences between trained models. It includes locate-then-edit methods for factual associations \citep{meng2022rome,meng2023memit}, learned editors \citep{decao2021editing,mitchell2022mend}, arithmetic on weight differences \citep{ilharco2023task}, and steering or ablation of directions in the residual stream \citep{turner2023actadd,li2023iti,arditi2024refusal}. Weight-space programming extends this family by giving a language model programmatic access to another network's weights and letting it decide, from measurements it requests, what to change \citep{takehi2026wsp}. Its first demonstration on Qwen3.5-0.8B \citep{qwen35card} removed a direction associated with repetitive looping from two MLP layers and raised MATH-500 accuracy \citep{hendrycks2021math,lightman2024verify} from 7.9\% to 48.1\% without a gradient step.

The two families have mostly been studied apart, and with different goals. Direct edits are usually evaluated on whether they change a fact or suppress a behaviour while leaving everything else untouched \citep{yao2023editing}; gradient methods are evaluated on how much of a new task they learn and, more recently, on how much they forget \citep{biderman2024lora,luo2025forgetting}. We compare closed-form weight editing with gradient training on the same backbone, spatial tasks and evaluation protocol, asking how each improves target performance and affects neighbouring skills.

We chose text-based spatial reasoning as the domain because it requires composing relations over several steps, a known difficulty for language models \citep{weston2016babi,shi2022stepgame,mirzaee2021spartqa}, and because its tasks have exact symbolic solutions against which a model's intermediate steps can be checked. We used four task families with two roles (Section~\ref{sec:tasks}). Two target tasks measure directional reasoning, on which the base model fails. The reference-frame task asks for the egocentric relation (ahead, behind, left, right) of a landmark given in map coordinates and an observer's heading, the conversion between allocentric and egocentric frames of reference that is central to human spatial language \citep{levinson2003space}. StepGame asks for the relation between two agents from relative-position statements at dataset difficulty levels 1 to 10 \citep{shi2022stepgame,li2024stepgame}. Two off-target families measure whether a change preserves a neighbouring skill that the patched model already has: comparing straight-line distances between points on a map, and the same comparison phrased in terms of product prices and weights, without spatial language. As the written edit we used \pub, the published weight-space-programming patch of Qwen3.5-0.8B, and as the trained alternative we used LoRA adapters of rank 1 and 16.

The comparison yields four results. First, the patch's gain on both tasks depends on the arrangement of its weight change, not on its magnitude, and five attempts to extend it with further closed-form edits did not improve on it. Second, LoRA with a rank-1 update, with a trainable parameter count of the same order as the closed-form edits, reached ceiling on the reference-frame task and near-ceiling on StepGame. Third, the trained adapters learned a procedure tied to the form of the training data: worked-solution supervision outperformed answer-only supervision, and transfer was stronger for new wording than for higher difficulty levels or unseen combinations of known elements. Fourth, the adapters applied that procedure to an untrained task that resembled their inputs and their accuracy collapsed, whereas the patch preserved performance.

\section{Background and Related Work}

Fine-tuning a pre-trained network changes behaviour reliably but can overwrite earlier competence, a problem first described for connectionist networks as catastrophic interference \citep{mccloskey1989} and addressed with regularisation \citep{kirkpatrick2017ewc} and rehearsal \citep{rolnick2019replay}. Low-rank adaptation adds a trainable product of two thin matrices to each frozen weight \citep{hu2022lora}. The observation that fine-tuning objectives can be optimised in low-dimensional subspaces \citep{li2018intrinsic,aghajanyan2021intrinsic} explains why very low ranks often suffice. Restricting the update does not remove forgetting: LoRA forgets less than full fine-tuning but also learns less \citep{biderman2024lora}, continual instruction tuning degrades general knowledge and reasoning \citep{luo2025forgetting}, fine-tuning can distort pretrained features and reduce out-of-distribution performance \citep{kumar2022distort}, and changes in implicit task inference can suppress existing capabilities \citep{kotha2024forgetting}.

Targeted editing draws on the computational structure of the network. Feed-forward layers act as key--value memories \citep{geva2021kv}, and individual neurons or low-rank directions can be associated with specific facts \citep{dai2022neurons,meng2022rome}. Locate-then-edit methods use this structure to solve for a closed-form update \citep{meng2022rome,meng2023memit}, learned editors predict the update from a gradient \citep{decao2021editing,mitchell2022mend}, and steering methods add or remove a direction in the activations at inference time \citep{turner2023actadd,li2023iti}. Removing a single direction can disable refusal \citep{arditi2024refusal}, which shows that some behaviours are mediated by very low-dimensional structure. Where an edit should be applied is not always where causal analysis locates the information \citep{hase2023localization}. Weight-space programming differs from these methods in that the edit is designed by a language model that requests measurements and writes code against the weights \citep{takehi2026wsp}. Its published results concern removing an obstacle, looping, that prevented a model from completing its reasoning.

Spatial question answering has been studied since the bAbI tasks \citep{weston2016babi}. StepGame added multi-hop chains and a large set of sentence templates \citep{shi2022stepgame}, SpartQA added more complex spatial descriptions \citep{mirzaee2021spartqa}, and a corrected release of StepGame removed label errors and showed that mapping each statement to coordinates makes the task largely mechanical \citep{li2024stepgame}; we use that release. Intermediate-step supervision and reasoning prompts, including scratchpads, chain-of-thought demonstrations and zero-shot reasoning instructions, help language models with multi-step computation \citep{nye2021scratchpad,wei2022cot,kojima2022zeroshot}.

Neural sequence models often fail to recombine known parts in new ways \citep{lake2018scan}, to extend procedures to longer inputs than those seen in training \citep{anil2022length}, or to compose subroutines reliably as depth grows \citep{dziri2023faith}. Written reasoning need not reflect the computation that produces the answer \citep{turpin2023unfaithful}, a point that becomes concrete in Section~\ref{sec:ood}. Linear probes \citep{alain2017probes,belinkov2022probing} are the usual way to ask whether a quantity is represented at a given layer, and we use them to decide whether a closed-form write has a signal to amplify.

\section{Experimental Setup}
\label{sec:setup}

\subsection{Models}

The editing and training experiments used Qwen3.5-0.8B at Hugging Face revision \texttt{2fc06364} as the base model \citep{qwen35card}, a hybrid architecture that combines attention with gated delta-rule layers \citep{yang2025gdn}. The weight-space-programming patch \pub\ (revision \texttt{d1671a17}) differs from the base model in exactly two tensors, the MLP down-projections of layers 15 and 19 (7{,}340{,}032 elements, of which 130{,}188 differ after BF16 storage); adding the float32 difference to the base weights reproduces \pub\ bit for bit. The project workstation is equipped with an Intel Core i9-14900K CPU and an NVIDIA RTX 4090 GPU.

\subsection{Tasks and Benchmarks}
\label{sec:tasks}

Qualitative spatial relations are conventionally divided into topological, directional and distance relations \citep{cohn2001qsr}, and directional relations are further distinguished by the frame of reference in which they are expressed \citep{levinson2003space}. The four task families of this study cover directional and distance relations and serve two roles (Table~\ref{tab:tasks}). All items come from one frozen task file and were used unchanged.

The two target tasks, on which the closed-form edits were designed and the adapters were trained, measure directional reasoning, and they were chosen to separate two component skills. The reference-frame task isolates a single transformation without chaining. A direction defined in a fixed, absolute frame (a map with north up) must be re-expressed in the relative frame defined by an observer's heading, which requires computing two coordinate differences, deciding which of them dominates, and rotating the resulting compass direction by the heading. StepGame measures the complementary skill. All of its relations are stated in one fixed frame, and the answer is obtained by composing relations along the query path, each expressed through one of many sentence templates (compass terms, clock positions, angles and plain prepositions), and the story also contains statements that do not lie on the path between the two queried agents. Both tasks have exact symbolic solutions against which the worked solutions of Section~\ref{sec:lora} and the models' written intermediate steps can be checked. On both, the base model has low accuracy and \pub\ answers correctly on only part of the items, so there is room for a change to show a gain.

The two off-target families measure what a change to the model should leave intact. The map family asks which of two points is nearer to a third in straight-line distance. It is a distance relation that uses the same coordinate format as the reference-frame task but a different computation, the comparison of $(\Delta x)^2+(\Delta y)^2$ between two candidates. Its 128 items were constructed so that simple shortcuts fail: in 64 of them the city-block distance $|\Delta x|+|\Delta y|$ gives the opposite answer to the Euclidean distance, in 64 the correct answer differs from the one obtained by comparing $x$ offsets alone, and each of the answers B and C occurs 64 times. The product family repeats every map item with the same numbers and the same answer but describes the points as products with a price and a weight, so it requires the same arithmetic without spatial language. A loss on both families indicates that the computation itself or its use has been disturbed; a loss on the map family alone would indicate a reaction to spatial wording. \pub\ answers more than 90\% of the items of both families correctly (Section~\ref{sec:offtarget}), so a drop on them is a cost of the change and not a pre-existing weakness.

\begin{table}[t]
\centering\footnotesize
\caption{The four task families. Examples are quoted verbatim from the frozen task file, with the line breaks of the StepGame story replaced by spaces and the closing answer-format instruction omitted (``Answer with one word.'', ``Return only the label.'' with the list of nine labels, or ``Answer with one letter.''). Target families were used for editing, training and the main evaluation; off-target families were used only to measure side effects (Section~\ref{sec:offtarget}).}
\label{tab:tasks}
\begin{tabularx}{\linewidth}{@{}>{\raggedright\arraybackslash}p{2.3cm}>{\raggedright\arraybackslash}p{3.4cm}>{\raggedright\arraybackslash}X>{\raggedright\arraybackslash}p{1.2cm}@{}}
\toprule
Family (role, items) & Spatial reasoning measured & Example item & Gold\\
\midrule
Reference frame (target, 128) & Directional relation; transformation from an absolute (map) to a relative (observer) frame; one step & \emph{On a flat map, x increases east and y increases north. An observer stands at (2, 87) facing west. A landmark is at (1, 62). Relative to the observer, is the landmark ahead, behind, left, or right?} & left\\
\addlinespace
StepGame (target, 200; 20 per level 1--10) & Directional relations in one fixed frame; composition along a query path; varied wording; irrelevant statements & \emph{X is on the right side to T. T is north east of W. C is over there and A is on the right. I and C are in a horizontal line with I on the left. I is on the left and X is on the right. What is the relation of the agent W to the agent I?} (level 5) & lower-left\\
\addlinespace
Map distance (off-target, 128) & Distance relation; comparison of Euclidean distances from coordinates & \emph{On a flat map, point A is at (3, 9), point B is at (75, 86), and point C is at (81, 8). Which point is closer to A in straight-line distance, B or C?} & C\\
\addlinespace
Product distance (off-target control, 128) & The same computation as map distance with no spatial wording; paired item by item with the map family & \emph{Product A has price 28 and weight 98, product B has price 3 and weight 22, and product C has price 23 and weight 6. Using Euclidean distance on (price, weight), which product is more similar to A, B or C?} & B\\
\bottomrule
\end{tabularx}
\end{table}

In the map example, comparing $x$ offsets alone (72 against 78) would select B, whereas the squared distances are 11{,}113 and 6{,}085, so the answer is C. In the StepGame example, placing $I$ at $(0,0)$ gives $X=(1,0)$ and $T=(0,0)$ from the fifth and first statements, and $W=T-(1,1)=(-1,-1)$ from the second, so $W$ is lower-left of $I$; the third and fourth statements are not needed.

The reference-frame task consists of 128 items of the form \emph{``On a flat map, x increases east and y increases north. An observer stands at $(o_x,o_y)$ facing $h$. A landmark is at $(t_x,t_y)$. Relative to the observer, is the landmark ahead, behind, left, or right?''} The items cover all 16 combinations of heading and answer, eight items each, and the offset along the minor axis is at most 0.4 times the offset along the major axis. With compass indices $c(\text{north})=0,\dots,c(\text{west})=3$ and relative indices $\rho(\text{ahead})=0,\ \rho(\text{right})=1,\ \rho(\text{behind})=2,\ \rho(\text{left})=3$, the gold answer is
\begin{equation}
\rho^{\ast} = \bigl(c(w)-c(h)\bigr) \bmod 4,\qquad
w=\begin{cases}\text{east if } \Delta x>0,\ \text{west otherwise}, & |\Delta x|>|\Delta y|,\\ \text{north if } \Delta y>0,\ \text{south otherwise}, & \text{otherwise,}\end{cases}
\label{eq:frame}
\end{equation}
where $\Delta x=t_x-o_x$ and $\Delta y=t_y-o_y$. The StepGame set consists of 200 items, 20 for each dataset difficulty level $k$ from 1 to 10, from the validation split of the corrected release (repository \nolinkurl{Fangjun-Li/SpatialLM-StepGame}, revision \texttt{1114cb58}). We use $k$ for the source dataset's level designation; the number of relations on the query path is determined by the story. Each statement relates two agents through one of nine labels, which correspond to the unit offsets $v\in\{-1,0,1\}^2$; for the relation of $P$ to $Q$ the gold label is $\bigl(\operatorname{sgn}(d_x),\operatorname{sgn}(d_y)\bigr)$ with $d=\mathrm{pos}(P)-\mathrm{pos}(Q)$ obtained by composing the offsets along the chain. For the two distance families the gold answer is the candidate $X\in\{B,C\}$ that minimises $(x_X-x_A)^2+(y_X-y_A)^2$; ties do not occur.

\subsection{Decoding and Scoring}

Items, prompts, system prompt, gold labels and scoring rules were fixed before the experiments reported here and verified by hash before every evaluation. The main comparisons used the chat template with thinking enabled, at most 32{,}768 new tokens, nucleus sampling \citep{holtzman2020nucleus} with temperature 0.6, top-$p$ 0.95 and top-$k$ 20, presence penalty 1.5 and batches of 64. A rule-based parser read the final label after the thinking block, and an output without a parseable label counts as incorrect. The base, \pub, prompting and principal LoRA comparisons used three seeds, $s_0+k$ for $k\in\{0,1,2\}$, where $s_0$ is 191071 for the reference-frame task and 201071 for StepGame, and each batch of 64 used seed $s_0+k+\text{offset}$. Single-seed editing and control evaluations are identified with their results; the thinking-off diagnostics are described in Section~\ref{sec:structure}.

\subsection{Evaluation Criteria and Statistical Tests}

A condition was counted as improved on a task when it exceeded \pub\ by at least ten percentage points. With $a_k$ the number of correct items for seed $k$ and $N$ the number of items, the criterion is
\begin{equation}
\tfrac13\textstyle\sum_{k} a_k \;\ge\; \bar a_{\pub} + 0.1N \quad\text{and}\quad \min_k a_k > \max_k a^{\pub}_k ,
\label{eq:criterion}
\end{equation}
which with $\bar a_{\pub}=28.3$ and $67.3$ requires a mean of at least 41.1/128 and 87.3/200 and every seed above 34 and 69. Paired comparisons between two models on the same items and seed used the exact two-sided McNemar test \citep{mcnemar1947}, $p = \min\{1,\ 2\sum_{i\le \min(b,c)}\binom{b+c}{i}2^{-(b+c)}\}$, where $b$ and $c$ count the items answered correctly by only one of the two models.

\section{Further Weight Edits Did Not Improve on the Public Patch}
\label{sec:closed}

Five experiments attempted to add to \pub\ with gradient-free edits of at most 0.1\% of the weights, and none improved on it. In the first, an independent editing model (GPT-6 Astra) was given tool access to the weights and activations of \pub\ and to a development subset of 32 reference-frame and 40 StepGame items. It submitted an edit to four MLP output matrices (85{,}519 BF16 values, 0.011\% of the weights) that reinforced a direction separating early and late states of failing trajectories. On the full benchmark this edit scored 31/128 and 65/200, below the seed-matched \pub\ scores of 34/128 and 69/200 and indistinguishable from a matched random perturbation (26/128 and 67/200). A second editing session targeted the inversion of a queried relation in StepGame; the selected activation direction worsened the log-probability margin of the correct continuation more than a random direction of the same norm, and no weight was written.

The remaining experiments were designed from mechanistic measurements. A mapping that depends on the heading, written as eight conjunctive neurons in layer 20, in the key--value form of an MLP layer \citep{geva2021kv}, raised the teacher-forced accuracy of the mapping statements for non-north headings on held-out trajectories from 0.267 to 0.653, short of the pre-set gate of 0.70. On the development subset this edit gave 10/32 on the reference frame (\pub\ 7, gate 11) and 13/40 on StepGame (\pub\ 17, gate 15), and it was not carried to the full benchmark. A final experiment looked for the answer itself at the conclusion positions of \pub's reasoning. A ridge-regularised linear probe \citep{alain2017probes} in layer 20 predicted the gold answer with accuracy 0.376 against a gate of 0.60, and the comparison of $|\Delta x|$ with $|\Delta y|$ that Equation~\ref{eq:frame} requires was readable with accuracy 0.536, close to the chance level of 0.5. The probe did not meet the selection gate, and no candidate was produced.

\section{The Public Patch's Gains Depended on Edit Structure}
\label{sec:structure}

Three simpler questions about \pub\ itself precede the comparison with training (Table~\ref{tab:main}). The first is whether its gain comes from the specific pattern of its edit or from perturbing two MLP matrices by a given amount. Let $\Delta_\ell = W^{\pub}_\ell - W^{\text{base}}_\ell$ for each of the two edited tensors. We formed $W'_\ell = W^{\text{base}}_\ell + \operatorname{vec}^{-1}\!\bigl(\Pi\,\operatorname{vec}(\Delta_\ell)\bigr)$, where $\Pi$ is a uniformly random permutation of the tensor's elements, which keeps the distribution, number of non-zero entries and norm of the change but destroys its arrangement. Two independent permutations scored 1/128 and 3/128 on the reference frame and 39/200 and 40/200 on StepGame, the same as the base model (0/128 on all three seeds; 39, 38 and 38/200) and far below \pub. The gain of \pub\ therefore depends on the structure of its edit. The base model's failures were almost entirely non-terminations: 126 to 127 of its 128 reference-frame outputs and 135 to 140 of its 200 StepGame outputs reached the 32{,}768-token limit, whereas \pub\ reached it on 3 and 5. The patch's effect on these tasks is thus consistent with the repair of looping reported for mathematics \citep{takehi2026wsp}.

The second question is whether the procedure that training later supplied could be supplied by the prompt instead, as instructions or demonstrations often can \citep{brown2020fewshot,kojima2022zeroshot}. We prefixed each original prompt with a description of the solution method, and in a second condition with the same description followed by two worked examples generated by the training programs of Section~\ref{sec:lora} (not benchmark items); the original prompt followed unchanged. Neither condition improved \pub. The method description gave 26.0/128 and 61.0/200, and the worked examples gave 10.0/128 and 65.3/200. Many of these outputs ended their reasoning without a parseable label, for example 96 to 103 of 128 reference-frame outputs in the worked-example condition, so the instructions changed the form of the answers more than their correctness.

The third question is what the model does when the thinking block, in which the base model loops, is switched off. This diagnostic locates the effect of the patch and is reported separately from Table~\ref{tab:main}. With the original prompts, which ask for one word or one label, the base model and \pub\ both answered in one or two tokens (one seed, at most 1{,}024 new tokens, a limit that no output reached) and scored at the level of always choosing the most frequent label (32/128 and 38/200): 36 and 38/128 on the reference frame and 38 and 36/200 on StepGame. Without a reasoning trace the patch therefore had no effect. We then gave the base model, still with thinking off, a system prompt that asks it to reason step by step before answering and to write the answer on a final line of the form \texttt{Answer: <label>}; the user prompts were unchanged, and decoding otherwise followed Section~\ref{sec:setup}, including the limit of 32{,}768 tokens. The model then wrote its reasoning in the visible answer, on average 1{,}300 to 1{,}800 tokens on the reference frame and 4{,}300 to 4{,}800 on StepGame, and reached the token limit on at most 2 and 10 items per run. It scored 38.0/128 (31, 40 and 43) on the reference frame and 51.0/200 (51, 52 and 50) on StepGame. On the two distance families, on which the base model scores 13 and 17 with thinking on because it does not finish (Section~\ref{sec:offtarget}), it scored 123.0/128 and 114.7/128, close to the 120.3 and 118.0 of \pub. Adding the method description to this prompt lowered the target scores to 31.3/128 and 42.3/200, and adding the worked examples gave 34.7/128 and 40.7/200.

When the base model was kept from looping, it therefore solved the distance questions as well as \pub\ and part of StepGame. On StepGame the seed-0 score exceeded every one of 10{,}000 random reassignments of its predictions that keep their label frequencies, but the mean stayed below the 67.3 of \pub. On the reference frame the mean remained close to the most-frequent-label level of 32, which \pub\ with thinking on (28.3) does not exceed either, and the seed-0 score was not distinguishable from such a reassignment ($p=0.32$). These results support the reading that the base model's failures in thinking mode were mainly non-terminations of an existing computation, while reference-frame performance remained near the most-frequent-label baseline.

\begin{table}[t]
\centering\small
\caption{Accuracy on the original benchmark items (correct items per seed and mean). Changed or trainable parameters are given as a fraction of the language-model parameters. \pub\ is the published weight-space-programming patch; ``further closed-form edit'' is the best submitted candidate of Section~\ref{sec:closed} (one seed). The shuffled-delta rows are two independent permutations, each evaluated with seed 0. An asterisk marks conditions that meet the improvement criterion of Equation~\ref{eq:criterion} on that task.}
\label{tab:main}
\begin{tabular}{lrcrcr}
\toprule
& & \multicolumn{2}{c}{Reference frame (/128)} & \multicolumn{2}{c}{StepGame (/200)}\\
\cmidrule(lr){3-4}\cmidrule(lr){5-6}
Condition & Changed & Seeds & Mean & Seeds & Mean\\
\midrule
Base model & -- & 0 / 0 / 0 & 0.0 & 39 / 38 / 38 & 38.3\\
Base + shuffled \pub\ delta & 0.016\% & 1, 3 & 2.0 & 39, 40 & 39.5\\
\pub & 0.016\% & 34 / 25 / 26 & 28.3 & 69 / 67 / 66 & 67.3\\
\pub\ + further closed-form edit & 0.011\% & 31 & 31.0 & 65 & 65.0\\
\pub\ + method instruction & -- & 24 / 26 / 28 & 26.0 & 56 / 61 / 66 & 61.0\\
\pub\ + worked examples & -- & 12 / 8 / 10 & 10.0 & 70 / 59 / 67 & 65.3\\
\midrule
LoRA rank 16, answer-only & 1.44\% & 32 / 26 / 33 & 30.3 & 29 / 17 / 27 & 24.3\\
LoRA rank 1 & 0.09\% & 128 / 128 / 128 & 128.0$^\ast$ & 196 / 198 / 199 & 197.7$^\ast$\\
LoRA rank 16 & 1.44\% & 128 / 128 / 127 & 127.7$^\ast$ & 191 / 186 / 190 & 189.0$^\ast$\\
LoRA rank 16, 1{,}792 examples & 1.44\% & 127 / 126 / 128 & 127.0$^\ast$ & 193 / 194 / 190 & 192.3$^\ast$\\
LoRA rank 16, 448 examples & 1.44\% & 123 / 123 / 122 & 122.7$^\ast$ & 185 / 180 / 179 & 181.3$^\ast$\\
LoRA rank 16, trained from base & 1.44\% & 120 / 121 / 119 & 120.0$^\ast$ & 194 / 195 / 193 & 194.0$^\ast$\\
\bottomrule
\end{tabular}
\end{table}

\section{LoRA Surpassed the Public Patch on Both Spatial Tasks}
\label{sec:lora}

\subsection{Training Procedure}

We trained LoRA adapters on \pub\ with supervision on a worked solution written inside the thinking block, in the manner of scratchpad supervision \citep{nye2021scratchpad}. Table~\ref{tab:setup} lists the settings. For the reference frame, 2{,}048 training scenes (128 for each heading--answer combination) and 128 held-out scenes were drawn from the benchmark's generator with a new seed, excluding every scene that appears in any task file of the project. The target text states the heading's mapping, computes the offsets, names the dominant world direction and converts it, for example
\begin{quote}\footnotesize\ttfamily
Facing north: ahead is north, right is east, behind is south, left is west.\\
dx = 8 - 10 = -2, dy = 74 - 49 = 25.\\
|dx| = 2 and |dy| = 25, so the y offset dominates and the landmark lies mainly north.\\
North is ahead for an observer facing north.
\end{quote}
followed by the closing thinking tag and the label. For StepGame, 512 training and 20 held-out problems per hop were drawn from the official training split. Problems that matched a benchmark item after renaming the agents in order of appearance (1{,}378, including the one exact duplicate) and problems with a story identical to a benchmark story were excluded. The worked solution was produced by a symbolic solver that maps each of 196 sentence templates, induced from one-hop training problems only, to an offset and composes the offsets along the chain (\texttt{Offsets (x right, y up): ... Put P at (0, 0) ...}). The solver reproduced the official label of all 200{,}265 training problems, all 6{,}639 validation problems and all 200 benchmark items; benchmark labels were used only for this check. No model output entered any training target.

With $W$ a frozen linear weight, the adapter adds $\Delta W = s\,BA$ with $A\in\mathbb{R}^{r\times d_{\text{in}}}$, $B\in\mathbb{R}^{d_{\text{out}}\times r}$ and scale $s=20$ \citep{hu2022lora}, applied to the linear modules within all 24 decoder layers. The loss is the token-mean cross-entropy over the target tokens $\mathcal{T}$ only (worked solution, closing tag, label and end-of-turn token),
\begin{equation}
\mathcal{L}(\theta) = -\frac{1}{\sum_{b}|\mathcal{T}_b|}\sum_{b}\sum_{t\in\mathcal{T}_b}\log p_\theta\bigl(y_{b,t}\mid y_{b,<t}\bigr).
\end{equation}
Each training batch of 16 was split into four micro-batches with per-layer gradient checkpointing \citep{chen2016checkpoint}; weighting each micro-batch by its share of target tokens makes the accumulated gradient identical to that of a single batch. Checkpoints were saved at 384, 768 and 1{,}536 steps and scored on the generator held-out sets. The first checkpoint that reached 0.90 on the reference frame and 0.50 on StepGame was selected, and benchmark items were never used for selection.

\begin{table}[t]
\centering\small
\caption{LoRA training settings shared by all gradient conditions unless a row of Table~\ref{tab:main} states otherwise.}
\label{tab:setup}
\begin{tabularx}{\linewidth}{lX}
\toprule
Starting weights & \pub\ (one condition: base model)\\
Adapter & rank 16 (one condition: rank 1), scale 20, dropout 0, all 24 layers; 10{,}822{,}656 trainable parameters (1.44\%), rank 1: 676{,}416 (0.090\%)\\
Data & 2{,}048 reference-frame scenes and 5{,}120 StepGame problems (512 per hop), shuffled together with seed 28 (7{,}168 examples)\\
Target & worked solution inside the thinking block, closing tag, label; loss on target tokens only\\
Optimiser & Adam \citep{kingma2015adam}, learning rate $10^{-4}$, batch 16 (4 micro-batches of 4 with gradient checkpointing), training seed 28\\
Selection & generator held-out sets (128 frame, 200 StepGame), seed 281001; gate 0.90 and 0.50; checkpoints 384, 768, 1{,}536\\
Evaluation & original items, three seeds, decoding as in Section~\ref{sec:setup}\\
\bottomrule
\end{tabularx}
\end{table}

\subsection{Target-Task Performance}

\begin{figure}[t]
\centering
\includegraphics[width=\linewidth]{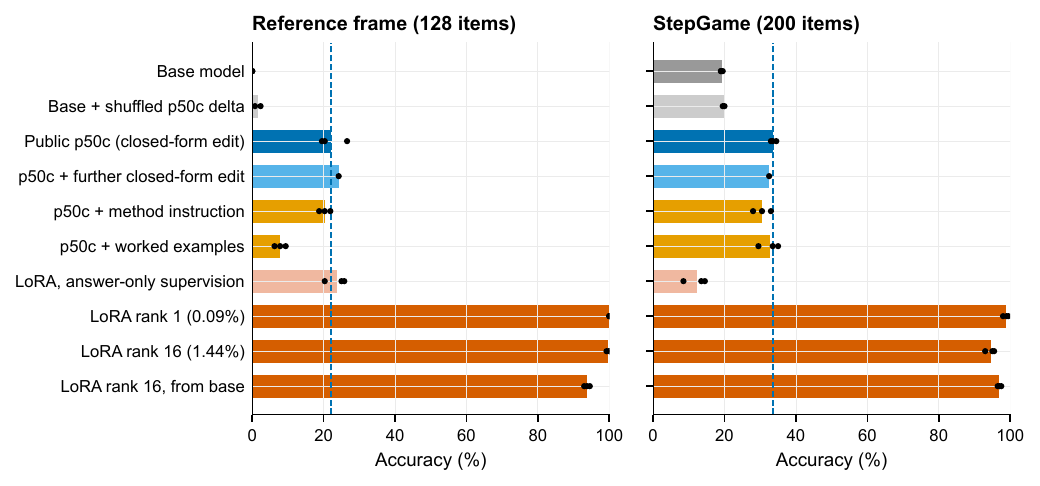}
\caption{Accuracy of the base model, the weight-space-programming patch \pub\ and its variants, prompting baselines and LoRA adapters on the original reference-frame (128 items) and StepGame (200 items) benchmarks. Bars show the mean over seeds and dots the individual seeds; the dashed line marks the three-seed mean of \pub. Randomly permuting the elements of \pub's weight change within each tensor returns accuracy to the base level, prompting does not improve on \pub, and LoRA adapters trained on worked solutions approach ceiling on both tasks even at rank 1, provided the supervision includes the intermediate steps.}
\label{fig:main}
\end{figure}

Gradient training succeeded where the closed-form edits had not (Figure~\ref{fig:main}). An adapter trained on the reference frame alone reached 128/128 on all three seeds, but it answered every StepGame question with the reference-frame template and fell to 12/200 on seed 0. Training on both tasks together removed this collapse: the mixed adapter passed the selection gate at the first checkpoint (384 steps, held-out 128/128 and 190/200) and scored 128, 128 and 127 on the reference frame and 191, 186 and 190 on StepGame, meeting Equation~\ref{eq:criterion} on both tasks. Its outputs were short and always terminated.

To exclude the possibility that the gain came from the short answer format rather than from what the supervision said, we trained a control adapter on the same data, in the same order and for the same 384 steps, with a different initialisation seed and with every conclusion and label shifted to a fixed wrong label (one step clockwise for the reference frame; the next label in the label list for StepGame). Scored against the true gold labels it obtained 0/128 and 0/200, and its predictions matched the shifted labels on 128 of 128 and 197 of 200 items. The adapters therefore carry out the mapping that they were taught.

\section{Supervision Mattered More Than Adapter Rank}
\label{sec:ablation}

Each of three ablations changed one property of the mixed adapter and kept everything else (Table~\ref{tab:main}). Reducing the rank from 16 to 1 reduced the trainable parameters to 676{,}416, or 0.09\% of the model, the same order of magnitude as the closed-form edits of Section~\ref{sec:closed}. The rank-1 adapter passed the gate at 384 steps and scored 128/128 on all seeds and 196, 198 and 199/200. This result is consistent with the low intrinsic dimension of fine-tuning objectives \citep{li2018intrinsic,aghajanyan2021intrinsic}: high target accuracy was achieved with a small trainable update.

\begin{figure}[t]
\centering
\includegraphics[width=\linewidth]{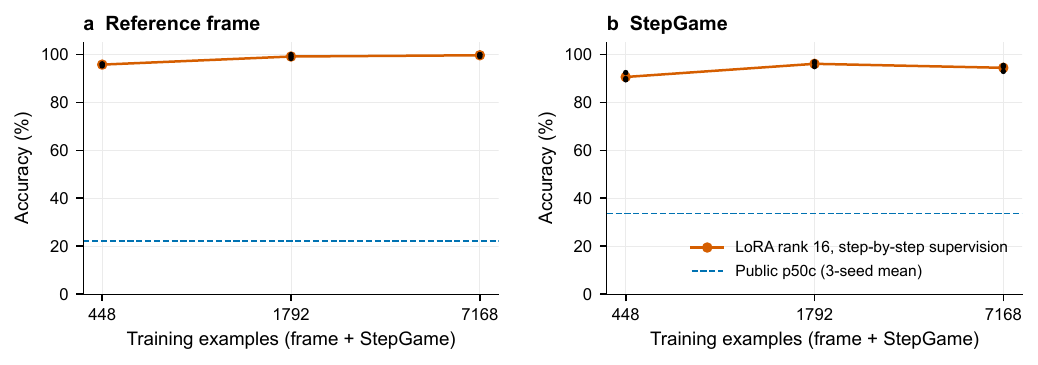}
\caption{Accuracy of rank-16 LoRA adapters trained with step-by-step supervision on 448, 1{,}792 and 7{,}168 examples (reference-frame scenes plus StepGame problems in a fixed 2:5 ratio), evaluated on the original benchmark items with three seeds (dots) and their mean (line). The dashed line is the three-seed mean of the weight-space-programming patch \pub. Accuracy is already far above \pub\ with 448 examples and saturates by 1{,}792.}
\label{fig:curve}
\end{figure}

Keeping the data but removing the worked solution from the target, so that the model was supervised only on the label after thinking freely, reduced target accuracy. No checkpoint passed the gate (the best, at 768 steps, reached 0.27 and 0.08 on the held-out sets), and the selected checkpoint scored 30.3/128 and 24.3/200, the latter well below \pub. Worked-solution supervision produced the target gains in this comparison, consistent with the value of intermediate steps for multi-step computation \citep{nye2021scratchpad,wei2022cot}. The amount of data needed was small (Figure~\ref{fig:curve}). Using the first 448 examples of the training file (128 reference-frame scenes and 32 StepGame problems per hop, about 14 epochs at 384 steps) already gave 122.7/128 and 181.3/200, and 1{,}792 examples matched the full set.

Finally, we trained the same adapter from the base model instead of \pub, with all other settings unchanged. It passed the gate at the same step and scored 120, 121 and 119/128 and 194, 195 and 193/200, with no output reaching the token limit although the base model itself exhausted the budget on almost every item. A margin of five percentage points around the adapter trained from \pub\ (121.3 and 179.0) was set in advance for concluding that the starting point does not matter. StepGame was within the margin and slightly higher; the reference frame was 7.7 items (6.0 points) below it, so by that rule \pub\ helped the training on the reference frame. Training does not need \pub\ to reach high accuracy on either task.

\section{Transfer Was Stronger to New Wording Than to\texorpdfstring{\\}{ }New Combinations}
\label{sec:ood}

\begin{figure}[t]
\centering
\includegraphics[width=\linewidth]{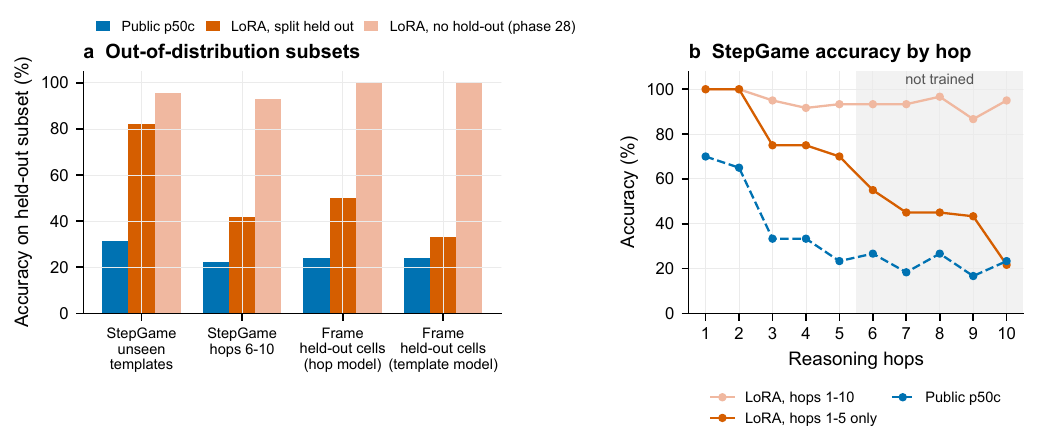}
\caption{Generalisation of LoRA adapters to parts of the input space removed from training, measured on the corresponding subsets of the original benchmark items (three-seed means). (a) Accuracy of \pub, the adapter trained without the held-out part, and the adapter trained on everything. (b) StepGame accuracy by dataset difficulty level for \pub, an adapter trained on all levels, and an adapter trained on levels 1--5 only; the shaded region contains levels absent from that adapter's training data. Accuracy is highest on unseen sentence templates and lower on held-out difficulty levels and heading--answer combinations.}
\label{fig:ood}
\end{figure}

The training data of Section~\ref{sec:lora} came from the same generator or the same corpus as the benchmark items. To measure generalisation without changing the benchmark, we trained two further mixed adapters with parts of the input space removed from training and scored them on the subsets of the original items that fall into the removed parts (Figure~\ref{fig:ood}, Table~\ref{tab:ood}). In both adapters the reference-frame training data excluded four heading--answer combinations, (north, ahead), (east, right), (south, behind) and (west, left), so that each heading and each answer was still seen with three other partners, a split in the spirit of compositional generalisation tests \citep{lake2018scan}. The ``hop'' adapter was trained on StepGame problems at levels 1 to 5 only (1{,}024 per level) and tested on the 100 benchmark items at levels 6 to 10, a test of transfer across dataset difficulty levels motivated by length-generalisation studies \citep{anil2022length}. The ``template'' adapter was trained on all levels but without 46 of the 196 sentence templates, chosen at random within each direction, and tested on the 140 benchmark items that contain at least one held-out template.

\begin{table}[t]
\centering\small
\caption{Accuracy on the held-out subsets of the original items (mean of three seeds). ``No hold-out'' is the mixed adapter of Section~\ref{sec:lora}, which saw all parts of the input space. The McNemar test compares the held-out adapter with \pub\ on seed 0 ($b$: only the adapter correct; $c$: only \pub\ correct).}
\label{tab:ood}
\begin{tabular}{lrrrrl}
\toprule
Held-out subset & Items & \pub & Held out & No hold-out & McNemar $b$/$c$, $p$\\
\midrule
StepGame, unseen templates & 140 & 0.31 & 0.82 & 0.95 & 76/5, $2\times10^{-17}$\\
StepGame, levels 6--10 & 100 & 0.22 & 0.42 & 0.93 & 26/5, $2\times10^{-4}$\\
Frame, unseen cells (hop adapter) & 32 & 0.24 & 0.50 & 1.00 & 13/7, 0.26\\
Frame, unseen cells (template adapter) & 32 & 0.24 & 0.33 & 1.00 & 9/8, 1.0\\
\bottomrule
\end{tabular}
\end{table}

Generalisation depended on what was held out. New wording transferred well: on items containing unseen templates the template adapter scored 0.82, close to the 0.95 of the adapter that had seen every template. Transfer to higher difficulty levels was partial: the hop adapter scored 0.42 at levels 6 to 10, twice the 0.22 of \pub\ but far below the 0.93 obtained when those levels were included in training, and its accuracy fell steadily with dataset level, from 0.55 at level six to 0.22 at level ten, consistent with increasing difficulty in compositional tasks \citep{dziri2023faith}. It was also lower than the full adapter at the trained levels 3 to 5 (0.75, 0.75 and 0.70 against 0.95, 0.92 and 0.93). Transfer to new combinations was weaker: on the four held-out heading--answer cells the two adapters scored 0.50 and 0.33, neither significantly different from \pub. Reading these outputs showed a specific failure. In the held-out cells whose world direction is south, (east, right) and (west, left), almost every answer was \emph{behind}, the answer that south has for a north-facing observer. For a scene with heading east, an adapter would correctly write ``right is south'' in the mapping line and then conclude ``South is behind for an observer facing east''. The correct mapping statement was followed by an incompatible conclusion, illustrating the inconsistency between written reasoning and final answers discussed by \citet{turpin2023unfaithful}. The outputs followed a familiar heading--answer association instead of the transformation in Equation~\ref{eq:frame}.

\section{Spatial Adaptation Disrupted Distance Reasoning}
\label{sec:offtarget}

The single-task adapter of Section~\ref{sec:lora} had shown that training on one task can take over another. To ask whether the mixed adapters had a similar side effect, we evaluated \pub\ and three adapters (rank 16 from \pub, rank 1, and rank 16 from base) on the two distance-comparison families that were part of the original task file but never used in training or selection. The first asks which of two points is closer to a third on a map; the second asks the same question about products described by price and weight. Seeds and decoding followed the original measurement of these families (seeds 171071 and 181071, plus $k$).

\begin{figure}[t]
\centering
\includegraphics[width=\linewidth]{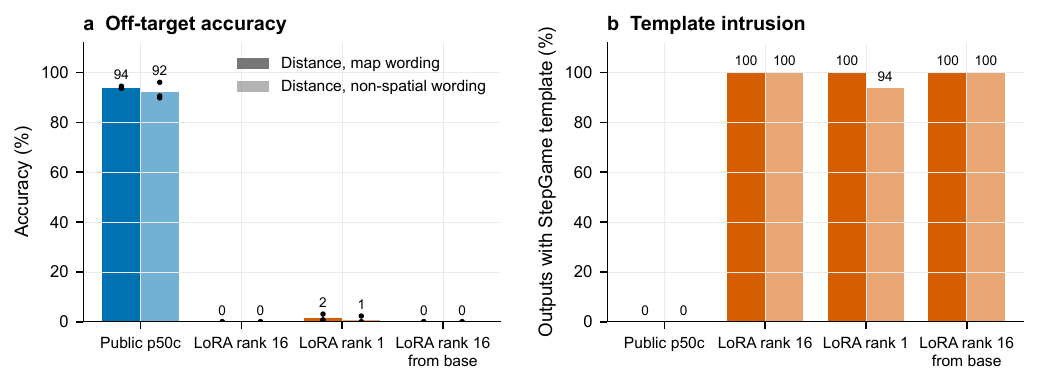}
\caption{Effect of training on two tasks outside the training data: comparing straight-line distances between points on a map, and the same comparison phrased in terms of product price and weight (128 original items each, three seeds). (a) Accuracy, with seed values as dots and the mean printed above each bar. (b) Percentage of outputs that contain the StepGame worked-solution template. The weight-space-programming patch \pub\ answers more than 90\% correctly and never uses the template; every LoRA adapter applies the StepGame template to almost every item and answers with a direction label, so its accuracy falls to between 0\% and 2\%.}
\label{fig:offtarget}
\end{figure}

The result was a near-complete collapse in distance-task accuracy (Figure~\ref{fig:offtarget}). \pub\ answered 120.3/128 and 118.0/128 correctly on average, whereas all three adapters scored between 0 and 2 on both families; the rank-16 adapters scored 0 on every seed (McNemar against \pub\ on seed 0, $p<10^{-34}$ for every adapter and family). The outputs showed a consistent change in task interpretation. The adapters treated the distance question as a StepGame problem: they opened the thinking block with \texttt{Offsets (x right, y up):}, converted the points into offsets, and after the thinking block returned a direction label such as \emph{upper-right} instead of the letter B or C that the question asked for. This happened on all 128 items of the map-worded family for every adapter and seed, and on all items of the product-worded family for both rank-16 adapters, although that family contains no spatial language. The rank-1 adapter entered the same template and then repeated its last steps until it reached the token limit on 121 to 128 items per run. In the seed-0 outputs, which we classified individually, the reference-frame template never appeared. The base model, for comparison, scored 13 and 17 on these families, mostly because it did not finish.

The adapters had therefore learned a procedure that is triggered by the presence of coordinates or paired numbers in a relational question and that displaced distance reasoning in its outputs. This matches the view that fine-tuning can change which task a model infers from its input, so that a capability is suppressed rather than erased \citep{kotha2024forgetting}. The closed-form patch, which improved the target tasks much less, left that reasoning intact.

\section{Discussion}

The experiments separate several questions that the comparison between writing and training weights tends to merge. A rank-1 adapter reached high target accuracy, showing that a small trainable update can support the task mapping. The contrast with unsuccessful closed-form edits directs attention to how the update is constructed. The closed-form routes tested here sought activation signals that a write could strengthen or remove. For the reference frame, the linear probe of the offset-magnitude comparison remained near chance at the inspected conclusion positions. Gradient training achieved high accuracy with worked solutions, whereas answer-only training failed to pass the learning gate. What it learned was a procedure tied to the form of its training data. It extended to new wording, degraded at higher difficulty levels, transferred weakly to new combinations, and was applied to every question that resembled its inputs, including questions to which it did not belong.

The weight-space-programming patch shows the opposite profile. Its gain on the spatial tasks is modest and consistent with removing the looping that kept the base model from finishing, an obstacle-removing edit of the kind that single-direction interventions achieve for refusal. That gain is tied to the arrangement of its weight change, and it costs nothing measurable on the neighbouring distance tasks. The choice between the two is therefore not between a method that works and one that does not. Training on worked solutions is the reliable way to install a specific procedure, but it should be expected to install that procedure wherever its trigger appears unless the training data say otherwise. A written edit is narrower in what it can currently achieve and gentler in what it disturbs. Whether the two can be combined, for example by using a trained adapter to locate a computation and a written edit to install it locally, is an open question that the present results make concrete.

We also asked whether the comparison could be carried to small instruction-tuned models that have no thinking mode, Gemma 3 1B (\nolinkurl{google/gemma-3-1b-it}) and Llama 3.2 1B (\nolinkurl{meta-llama/Llama-3.2-1B-Instruct}). They were evaluated on the same 584 items with the same decoding settings, with thinking necessarily off. With the original prompts both answered the target tasks in one to three tokens, at or below the level of always choosing the most frequent label (32/128 on the reference frame and 38/200 on StepGame). We then used the step-by-step system prompt of Section~\ref{sec:structure}, and for the two target tasks added, in two further conditions, the method description and the worked examples used with \pub. No condition lifted either model above the most-frequent-label level on either target task; three-seed means ranged from 0.3 to 30.7 of 128 on the reference frame and from 14.7 to 28.3 of 200 on StepGame. The two models failed in different ways. Gemma wrote about 300 tokens of reasoning per item, but its arithmetic comparisons were wrong, for instance treating an offset of 1 as larger than one of 25, and with the method description it stopped at a compass direction instead of converting it into an observer-relative label. Llama ignored the instruction on the target tasks and still answered in one or two tokens; the method description and the worked examples made it write reasoning without raising its accuracy. On the distance families Gemma stayed near the chance level of 64/128, and most of Llama's outputs contained no parseable letter. These prompting runs did not establish the target or neighbouring-task performance needed for the same editing-versus-training comparison. The contrast with Qwen3.5-0.8B, which with the same prompt and thinking off solved the distance questions and part of StepGame (Section~\ref{sec:structure}), suggests that the edit studied here removes an obstacle to a computation the model already performs, while supervised adaptation supplies a route to high target accuracy.

The comparison covers one 0.8B model and two-dimensional directional and distance tasks; the two additional 1B models were tested with prompting only. Each adapter was trained once, so the three evaluation seeds measure sampling variation rather than training variation. Editing and LoRA also used different data volumes and computational budgets, and their parameter counts describe different update structures. The results therefore compare achieved behaviour rather than equal-budget efficiency. The unsuccessful edits and answer-only ablation establish outcomes for the tested methods, without making worked solutions universally necessary or ruling out other editing strategies. Similarly, probe failures do not establish the absence of a computation, and distance-task failures do not establish that its underlying capability was erased. StepGame's source-dataset levels do not directly measure query-path length, so the held-out-level comparison combines changes in story complexity and reasoning demands. All benchmark items had been inspected before these experiments; Section~\ref{sec:ood} provides the generalisation tests. Off-target evaluation covers two related task families, with individual classification of the absence of the reference-frame template performed for seed 0. Broader capability preservation and the effect of mixing tasks during training remain to be measured.

The most direct next step is to mix a small amount of off-target data into training, as rehearsal methods do for continual learning, and ask whether the take-over disappears while the target gain remains, together with repeated training runs for the main adapters. A second direction is to compare the mechanisms: the difference between \pub's localised change and the adapters' broad one should be visible in which inputs activate the changed components, and the ``written but unused'' mapping of Section~\ref{sec:ood} is a concrete target for causal analysis with probes and activation patching.

\bibliographystyle{unsrtnat}
\bibliography{refs}
\end{document}